%% file: main.tex
\documentclass[10pt,twocolumn,letterpaper]{article}

\usepackage{cvpr}              
\usepackage[table]{xcolor}
\usepackage{multirow}
\usepackage{indentfirst}
\usepackage{amsmath}
\usepackage{graphicx}

\usepackage{graphicx}
\usepackage{amsmath}
\usepackage{amssymb}
\usepackage{booktabs}
\usepackage{cuted}
\usepackage{caption}
\usepackage{cuted}
\usepackage{float}
\usepackage{graphicx}
\usepackage{pifont}
\usepackage{xcolor}
\usepackage{listings}
\usepackage{bm}

\definecolor{cvprblue}{rgb}{0.21,0.49,0.74}
\usepackage[pagebackref,breaklinks,colorlinks,citecolor=cvprblue]{hyperref}

\def\paperID{13559} 
\def\confName{CVPR}
\def\confYear{2024}

\title{GaussianShader: 3D Gaussian Splatting with Shading Functions\\ for Reflective Surfaces}

\author{Yingwenqi Jiang$^{1}$ \quad Jiadong Tu$^{1}$ \quad Yuan Liu$^{2}$ \quad Xifeng Gao$^{3}$\\ \quad Xiaoxiao Long$^{2,*}$ \quad Wenping Wang$^4$ \quad Yuexin Ma$^{1,*}$ \vspace{0.3em} \\
{\normalsize $^1$ShanghaiTech University} \quad
{\normalsize $^2$The University of Hong Kong} \quad 
{\normalsize $^3$Tencent America} \quad 
{\normalsize $^4$Texas A\&M University} \quad
}

\begin{document}
\maketitle

\begin{strip}
    \centering
    \vspace{-5em}
    \includegraphics[width=1.0\textwidth]{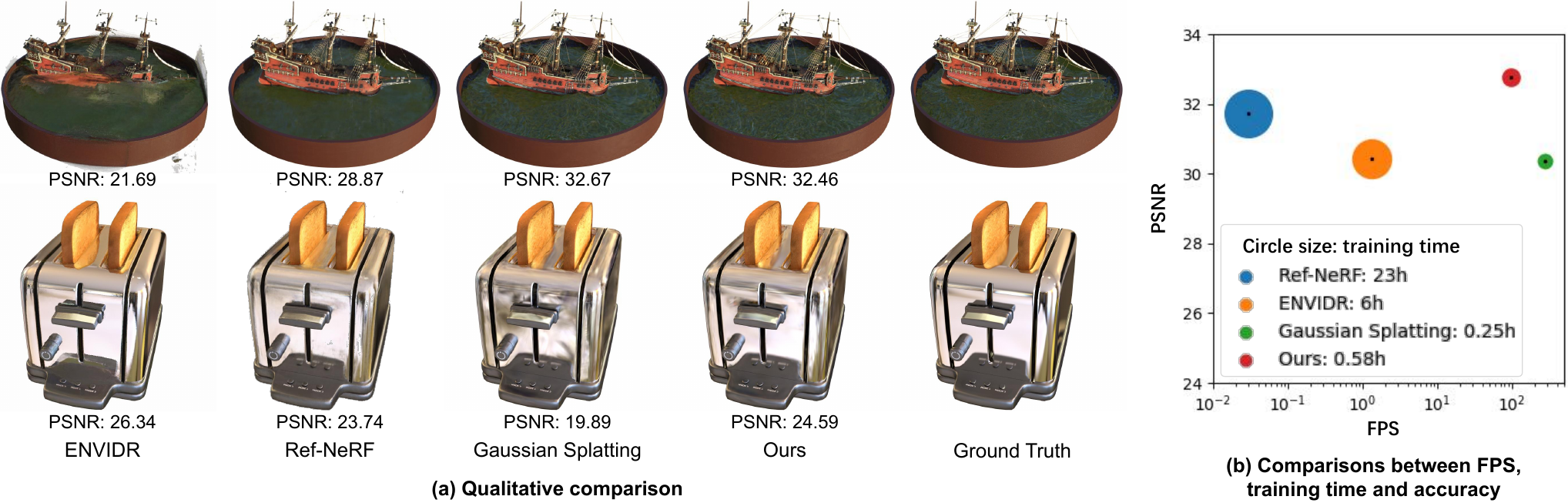}
    \vspace{-7mm}
    \captionof{figure}{
    GaussianShader maintains real-time rendering speed and renders high-fidelity images for both general and reflective surfaces. 
    Ref-NeRF\cite{verbin2022ref} and ENVIDR\cite{liang2023envidr} attempt to handle reflective surfaces, but they suffer from quite time-consuming optimization and slow rendering speed.
    3D Gaussian splatting~\cite{kerbl20233d} keeps high efficiency but cannot handle such reflective surfaces.
    }
    \label{fig:fig_teaser}
    \vspace{-3mm}
\end{strip}

\input{sec/0_abstract}    
\input{sec/1_intro}
\input{sec/2_related}
\input{sec/3_method}
\input{sec/4_experiment}
\input{sec/5_conclusion}
{
    \small
    \bibliographystyle{ieeenat_fullname}
    \bibliography{main}
}

WARNING: do not forget to delete the supplementary pages from your submission 
\input{sec/X_suppl}

\end{document}

%% file: sec/0_abstract.tex
\begin{abstract}
\vspace{-4mm}
The advent of neural 3D Gaussians~\cite{kerbl20233d} has recently brought about a revolution in the field of neural rendering, facilitating the generation of high-quality renderings at real-time speeds. However, the explicit and discrete representation encounters challenges when applied to scenes featuring reflective surfaces. 
In this paper, we present \textbf{GaussianShader}, a novel method that applies a simplified shading function on 3D Gaussians to enhance the neural rendering in scenes with reflective surfaces while preserving the training and rendering efficiency. 
The main challenge in applying the shading function lies in the accurate normal estimation on discrete 3D Gaussians. 
Specifically, we proposed a novel normal estimation framework based on the shortest axis directions of 3D Gaussians with a delicately designed loss to make the consistency between the normals and the geometries of Gaussian spheres.
Experiments show that GaussianShader strikes a commendable balance between efficiency and visual quality. Our method surpasses Gaussian Splatting~\cite{kerbl20233d} in PSNR on specular object datasets, exhibiting an improvement of 1.57dB. When compared to prior works handling reflective surfaces, such as Ref-NeRF~\cite{verbin2022ref}, our optimization time is significantly accelerated (23h vs. 0.58h). Please click on our \href{https://asparagus15.github.io/GaussianShader.github.io/}{project website} to see more results.
\end{abstract}

%% file: sec/1_intro.tex
\section{Introduction}
\label{sec:intro}
In recent years, the field of 3D computer vision has witnessed remarkable advancements in the 3D reconstruction and visualization of 3D scenes. Innovations such as Neural Radiance Fields (NeRF)~\cite{mildenhall2021nerf} have achieved substantial breakthroughs in generating novel views of 3D objects and scenes, presenting the potential for high-quality and photo-realistic renderings.
Despite these advancements, NeRF-related methodologies~\cite{barron2022mip, verbin2022ref, liang2023envidr},  still contend with challenges such as computationally expensive optimization and slow rendering speed. These limitations restrict their application in real-time interactive scenarios.

More recently, 3D Gaussian Splatting~\cite{kerbl20233d} combines 3D Gaussian representation and tile-based splatting techniques to achieve high-quality 3D scene modeling and real-time rendering, making it possible to employ neural rendering techniques in real applications. 
However, it suffers from a performance drop on scenes featuring specular and reflective surfaces. This is because 3D Gaussian Splatting~\cite{kerbl20233d} does not explicitly model appearance properties, so that fails to capture significant view-dependent changes, particularly specular highlights.
This constraint presents a substantial obstacle in the pursuit of achieving photorealistic rendering across a diverse array of materials, particularly those characterized by prominent reflective attributes.

Accurately modeling reflective surfaces is a challenging task. Ref-NeRF~\cite{verbin2022ref} and ENVIDR~\cite{liang2023envidr} combine the shading functions in implicit representations and present promising quality on reflective surfaces.
However, they suffer from time-consuming optimization (hours) and slow rendering speed. 
Due to the limited flexibility of SDF, ENVIDR~\cite{liang2023envidr} even fails to model complex scenes and presents a significant performance drop on general objects. It is still an unexplored problem how to combine the shading functions in a 3D Gaussian Splatting framework to improve its ability to handle reflections while preserving the efficiency in training and rendering.

In this paper, we present GaussianShader, a novel method that enhances the neural rendering of 3D Gaussians within scenes that contain reflective surfaces by incorporating a shading function on 3D Gaussians.
To ensure the efficiency of GaussianShader, evaluating the shading function cannot be too expensive while still retaining the ability to model the reflections. In light of this, we propose a novel simplified shading function that considers the diffuse colors and the direct reflections while putting all advanced complex reflections into a residual color term. In comparison with the shading function of Ref-NeRF which can only consider direct reflections, the utilization of this residual color enables GaussianShader to render more complex reflective appearances with efficiency.

Another challenge in computing a shading function is how to predict accurate normals on the discrete 3D Gaussian spheres. First, it is hard to get a locally-continuous surface from the 3D Gaussians to compute the surface normals. Second, associating multiple 3D Gaussians for normal computation would be very expensive using neighborhood searching. 
In GaussianShader, we address this problem by introducing a new normal representation, which is based on the shortest axis direction of a Gaussian sphere and learns a normal residual on this axis direction. Then, to enforce consistency between the estimated normals and the geometry formulated by Gaussian spheres, we introduce an efficient normal-geometry constraint between the predicted normals and the normals derived from the rendered depths.
Both the normal representation and the constraint lead to an accurate normal estimation of Gaussian spheres, which helps us compute the shading function.


Built upon 3D Gaussian Splatting, GaussianShader maintains real-time rendering speed while still accommodating various materials like reflective surfaces.
Experiments show that, compared to prior works, our method keeps a good balance between efficiency and robust performance on both general scenes and reflective surfaces. 

\textcolor{black}{
In summary, our proposed method offers several significant advantages:
\begin{enumerate}
    \item Our method explicitly approximates the rendering equation by a simplified shading function, significantly enhancing the realism of rendered scenes, particularly for highly specular and reflective surfaces.
    \item We propose a new normal estimation framework on 3D Gaussians with a new regularization loss that allows precise normal estimation.
    \item Leveraging the efficiency of Gaussian Splatting, our approach provides real-time rendering capabilities, making it suitable for interactive applications and scenarios that demand efficient rendering.
\end{enumerate}
}


%% file: sec/2_related.tex
\begin{figure*}[h]
    \centering
    \includegraphics[width=1.0\linewidth]{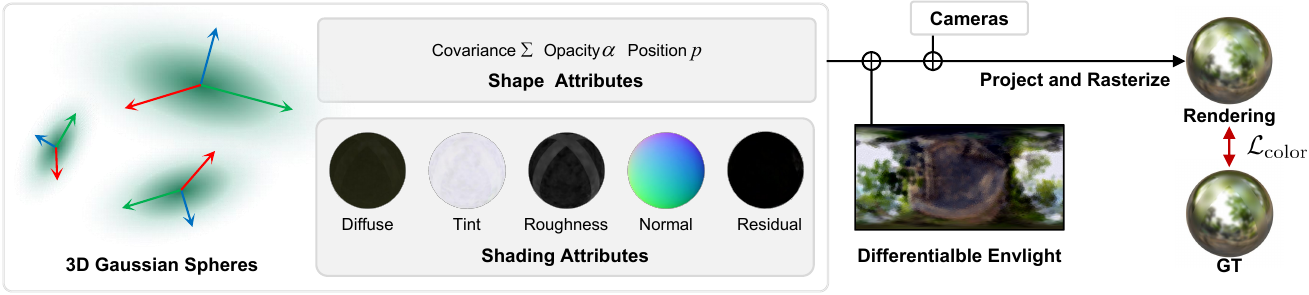}
    \vspace{-1ex}
    \caption{GaussianShader initiates with the neural 3D Gaussian spheres that integrate both conventional attributes and the newly introduced shading attributes to accurately capture view-dependent appearances. We incorporate a differentiable environment lighting map to simulate realistic lighting.
    The end-to-end training leads to a model that reconstructs both reflective and diffuse surfaces, achieving high material and lighting fidelity.}
    \label{fig:fig_pipeline}
    \vspace{-2ex}
\end{figure*}
\section{Related Work}
\subsection{Neural Radiance Fields}
Neural Radiance Fields (NeRF)~\cite{mildenhall2021nerf} gains remarkable progress in photo-realistic novel view synthesis using implicit representation and volume rendering. Recently, NeRF has inspired many follow-up works in various directions. \cite{barron2021mip, barron2022mip} improves NeRF in rendering quality by introducing 3D conical frustum, achieving state-of-art performance in NVS.  \cite{oechsle2021unisurf, yariv2021volume, wang2021neus, guo2022neural, yu2022monosdf} combine implicit surface representations with NeRF for more accurate geometric reconstruction. \cite{tancik2022block, turki2022mega} propose solutions for city-scale scene rendering using NeRF. \cite{srinivasan2021nerv, zhang2021nerfactor, yao2022neilf, verbin2022ref} targets a special kind of scenes with high specularities and reflections. Another important line of works \cite{muller2022instant, fridovich2022plenoxels, liu2020neural, chen2022tensorf, garbin2021fastnerf,  sun2022direct} focuses on acceleration due to the low training and rendering speed of NeRF using voxel grid or hash table. Although great progress has been made, NeRF-based methods still suffer from low rendering speed and high training-time memory usage due to their implicit nature \cite{chen2023text}. Based on works on point-based neural rendering \cite{zhang2022differentiable, xu2022point, ruckert2022adop, aliev2020neural}, a recent milestone work \cite{kerbl20233d} introduces anisotropic 3D Gaussian as an effective representation of the scene, and renders the image using a fast tile-based differentiable rasterizer, surpassing existing implicit neural representation methods in both quality and efficiency. 

\subsection{Reflective Object Rendering }

Rendering views of reflective objects from multi-view images has been a challenging task due to the complex light interactions. Previous approaches rely on simple light field interpolations to achieve high-fidelity rendering of novel perspectives~\cite{gortler2023lumigraph, levoy2023light, wood2023surface}, yet were constrained by the necessity for dense discrete captures. The accurate rendering of reflective surfaces hinges upon the precise estimation of scene  illumination (e.g. environment light) and material properties (e.g. BRDF), which is the task of inverse rendering~\cite{barron2014shape, nimier2019mitsuba}. Previous studies have demonstrated methodologies for predicting BRDF under known lighting conditions~\cite{deschaintre2018single, matusik2003data, aittala2016reflectance}, as well as techniques for estimating lighting given known geometry~\cite{legendre2019deeplight, richter2016instant, park2020seeing}.
Furthermore, \cite{zhang2022modeling, hasselgren2022shape, deng2022dip, yao2022neilf} incorporate indirect illumination, thereby enhancing the fidelity of the estimated BRDF.

Some NeRF related works attempt to model reflectance by disentangling the visual appearance into lighting and material properties, such as \cite{bi2020neural, zhang2021physg, boss2021nerd, srinivasan2021nerv, boss2021neural, zhang2021nerfactor}, which can jointly predict environmental illumination and surface reflectance properties under unknown or varying lighting conditions. Ref-NeRF~\cite{verbin2022ref} introduces a new parameterization and structuring of view-dependent outgoing radiance, as well as a regularizer on normal vectors.  Recent works~\cite{liang2022spidr, liu2023nero, liang2023envidr} utilize SDF-based representation to learn geometry from high specular surfaces, obtaining more accurate normals for physically based rendering. 
However, these method suffer from extremely time-consuming optimization and slow rendering speed, which hinders their employments in real applications.

%% file: sec/3_method.tex
\subsection{Preliminaries} 
\subsubsection{3D Gaussian Splatting Rasterization}
Our method builds upon Gaussian Splatting~\cite{kerbl20233d}, which begins with a collection of images capturing a static scene, their corresponding camera parameters, and a sparse point cloud generated through Structure-from-Motion (SfM)~\cite{snavely2006photo}. These points construct a set of Gaussians, each defined by position (mean) $\mathbf{p}$ and a 3D covariance matrix $\Sigma$.
While a direct optimization of the covariance matrix $\Sigma$ might seem intuitive, it presents challenges due to the requirement of positive semi-definiteness. As an alternative, in Gaussian Splatting, a more intuitive yet equally expressive representation is adopted as an ellipsoid for optimization. The ellipsoid is separated into scaling and rotation components, represented by a scaling matrix $S$ and a rotation matrix $R$, i.e., $\Sigma = RSS^TR^T$.

\subsubsection{Rendering with 3D Gaussians}
By projecting to 2D frame space according to the camera parameters, 3D Gaussian spheres are both differentiable and amenable to rapid rendering through 2D splatting with $\alpha$-blending. For radiance field modeling, the directional appearance aspect (color) $\mathbf{c}$ is conveyed through spherical harmonics (SH). By tile-based rasterization, we could sum up the pixel color $\mathbf{C}$ after sorting $\mathbf{c}_i$ by depth:
\begin{equation}\label{eq_tasterization}
    \mathbf{C} = \sum_{i \in N} \mathbf{c}_i \alpha_i \prod_{j=1}^{i-1} (1 - \alpha_j)
\end{equation}
where $\alpha_i$ is obtained by multiplying Gaussian weight with opacity $\alpha$ associated to Gaussian sphere.
After rasterization, the color loss could be applied to the rendered image:
\begin{equation}
\mathcal{L}_\text{color} = ||\mathbf{C}-\mathbf{C}_\text{gt}||^2
\end{equation}

\section{Method}
The overview of our method is depicted in Fig.~\ref{fig:fig_pipeline}. Our approach begins by adopting 3D Gaussian spheres, encompassing shape attributes including covariance $\Sigma$, opacity $\alpha$, and position $\mathbf{p}$. To enhance the representation ability on reflections, we compute the appearances of these Gaussian spheres with a shading function, which requires a set of shading attributes, including diffuse color, roughness, specular tint, normal, and residual color, as detailed in Sec.~\ref{sec_parameterization}.
Subsequently, we employ a differentiable environment light map to model the direct lighting, as elaborated in Sec.~\ref{sec_light}. The shading process relies significantly on accurate normal estimation, as discussed in Sec.~\ref{sec_normal}. 
Finally, we introduce the losses used in the whole training process in Sec.~\ref{sec_loss}.
\subsection{Shading on 3D Gaussians}\label{sec_parameterization}
Gaussian Splatting~\cite{kerbl20233d} models the appearances of Gaussians with simple spherical harmonic functions without considering the light-surface interactions. Thus, Gaussian Splatting fails to accurately represent strong specular surfaces.
However, accurately considering the light-surface interactions requires an exact evaluation of the Rendering Equation~\cite{kajiya1986rendering}, which requires extensive computational time and complex BRDF parameters.
We adopt a simplified approximation of the rendering equation which enables us to achieve high-quality rendering results on reflective surfaces in a considerably shorter time.

Specifically, for a Gaussian sphere, its rendered color $\mathbf{c}$ for the viewing direction $\omega_o$ is computed by
\begin{equation}
\mathbf{c}(\omega_o) = \gamma (\mathbf{c}_d + \mathbf{s} \odot L_s (\omega_o,\mathbf{n},\rho) + \mathbf{c}_r(\omega_o)),
\label{eq:shading}
\end{equation}
where $\gamma$ is a gamma tone mapping function~\cite{anderson1996proposal}, $\mathbf{c}_d\in [0,1]^3$ is the diffuse color of this Gaussian sphere, $\mathbf{s}\in [0,1]^3$ is the specular tint defined on this sphere, $L_s(\omega_o,\mathbf{n},\rho)$ is the direct specular light for this sphere in this direction, $\mathbf{n}$ is the normal of this Gaussian sphere, $\rho\in [0,1]$ is the roughness of the sphere, $\mathbf{c}_r: \mathbb{R}^3\to \mathbb{R}^3$ is so-called residual colors, and $\odot$ is the element-wise multiplication. 

\textbf{Explanations on Eq.~\ref{eq:shading}}. We explain our motivation of this shading model in the following three aspects.
\textbf{a}) Diffuse color $\mathbf{c}_d$ represents the consistent colors of this Gaussian sphere, which do not change with viewing directions.
\textbf{b}) $\mathbf{s} \odot L_s (\omega_o,\mathbf{n},\rho)$ describes the interactions between the surface intrinsic color $\mathbf{s}$ and the direct specular light $L_s$. This term enables us to represent most of the reflections in rendering.
\textbf{c}) Since there are still some reflections that cannot be explained by the above reflections of direct lights, such as scattering and reflection on indirect lights, we add a residual color term $\mathbf{c}_r(\omega_o)$ to account for these complex reflections. In comparison, Ref-NeRF \cite{verbin2022ref} adopts a similar shading function without such a residual color term, which makes it struggle to handle advanced complex reflections. $\mathbf{c}_r(\omega_o)$ is parameterized by spherical harmonic functions.
\textbf{d}) $\mathbf{c}_d$, $\mathbf{s}$, $\rho$, and the coefficients of the spherical harmonics in $\mathbf{c}_r(\omega_o)$ all are trainable parameters associated with this Gaussian sphere. In the following, we will introduce how to compute $L_s(\omega_o,\mathbf{n},\rho)$, and $\mathbf{n}$.

\begin{figure}
    \centering
    \includegraphics[width=0.7\linewidth]{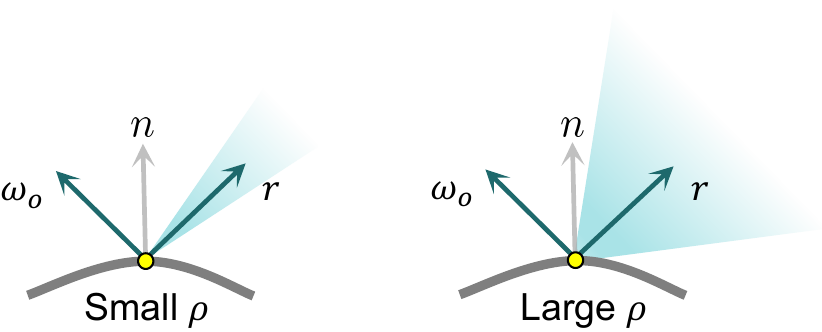}
    \caption{Normal Distribution Function D in Eq.~\ref{eq_light} is determined by the roughness $\rho$ and reflective direction $\mathbf{r}$. Surface with small $\rho$ has a smaller specular lobe and that with large $\rho$ has a larger specular lobe.}
    \label{fig:lobe}
      \vspace{-3ex}
\end{figure}
\subsection{Specular Light}\label{sec_light}
We compute the specular light $L_s$ by integrating incoming radiance with the specular GGX~\cite{walter2007microfacet} Normal Distribution Function $D$ visualized in Fig.~\ref{fig:lobe}
\begin{equation}\label{eq_light}
  L_s(\omega_o,\mathbf{n},\rho)=\int_{\Omega} L(\omega_i) D(\mathbf{r},\rho) (\omega_i \cdot \mathbf{n}) d\omega_i,
\end{equation}
where $\Omega$ represents the whole upper semi-sphere, $\omega_i$ is the direction for the input radiance and $D$ characterizes the specular lobe (effective integral range). When the surface is rough, the specular lobe will be larger around the reflective direction $\mathbf{r}$ while if the surface is smooth, the specular lobe will be smaller. The reflection direction $\mathbf{r}$ is calculated by view direction $\omega_o$ and normal $\mathbf{n}$ using $\mathbf{r} = 2({\omega}_o \cdot \mathbf{n})\mathbf{n} - {\omega}_o$
In our approach, environment light $L(\omega_i)$ is represented by a trainable $6\times64\times64$ cube map. 

This light integration is pre-filtered into multiple mip maps, where each mip map contains light integral of different reflective directions and different roughness. To compute the light integral for a specific roughness and a specific reflective direction, we only need to interpolate its values on the mip maps. In comparison to Ref-NeRF~\cite{verbin2022ref} which computes the light integral with integrated directional encoding, we choose the mip maps here because we find that these mip map-based light representations are more efficient in training. 
To get the reflective direction $\mathbf{r}$ for an observation direction $\omega_o$, we still need to estimate a normal $\mathbf{n}$ on a Gaussian sphere, which is introduced as follows.

\subsection{Normal Estimation}\label{sec_normal}
Normal estimation on Gaussian spheres is difficult. Gaussian spheres are a collection of discrete entities, each representing a localized point in space without a continuous surface or defined edge. This discrete structure makes it inherently difficult to directly calculate normals, which typically requires a continuous surface. In the following, we observed that the shortest axis direction of a Gaussian can serve as an approximated normal and we further associate a predicted normal residual on it.

\begin{figure}
    \centering
    \includegraphics[width=1.0\linewidth]{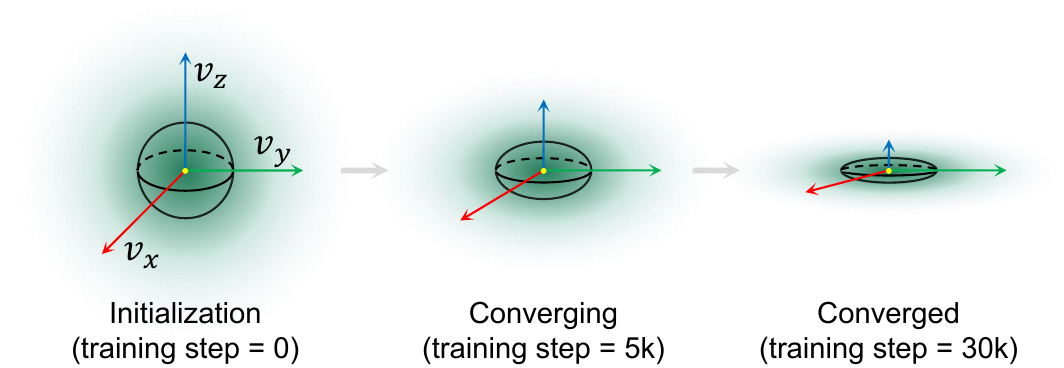}
    \caption{The geometric evolving process of a 3D Gaussian sphere in the optimization, which gradually becomes planar.}
    \label{fig:axis}
    \vspace{-3ex}
\end{figure}

\textbf{Shortest axis direction.}
In experiments, we made an interesting observation that the aspect ratio of the 3D Gaussian sphere—specifically, the ratios of their longest, intermediate, and shortest axis—gradually increased during optimization, as shown in Fig.~\ref{fig:axis}. This observation suggests that Gaussian spheres gradually become more flattened and approach a planar shape. This observation inspires us to select the shortest axis as the normal of this ``planarized" Gaussian sphere, denoted by $\mathbf{v}$.

\begin{figure}
    \centering
    \includegraphics[width=1.0\linewidth]{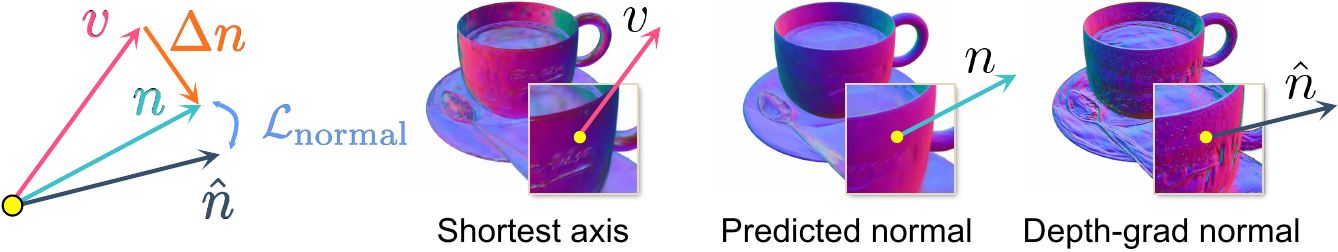}
    \caption{Visualization of the relationship between shortes axis $\mathbf{v}$, normal residual $\Delta \mathbf{n}$, normal $\mathbf{n}$ and depth-grad normal $\hat{\mathbf{n}}$. The supervision $\mathcal{L}_{normal}$ enforces the normal-geometry consistency.}
    \label{fig:normal}
      \vspace{-3ex}
\end{figure}

\textbf{Predicted normal residual.} 
The shortest axis $\mathbf{v}$ only serves as an approximated normal. To make the normal computation more accurate, we further introduce a trainable normal residual $\Delta \mathbf{n}$ on every Gaussian sphere.
However, the orientation of the shortest axis $\mathbf{v}$ has an ambiguity because the direction of the shortest axis could either point outward or inward from the surface. To handle this ambiguity, we optimize two separate normal residuals to accommodate both scenarios. Given a specific viewing direction $\omega_o$, we first select the direction aligned with the viewing direction $\mathbf{\omega_o}$ as the active normal direction for this viewing direction and then apply the corresponding normal residual to the active normal. This process is described by
\begin{equation}
    \mathbf{n} = \begin{cases}
          \mathbf{v} + \Delta \mathbf{n}_1 & \text{if $\mathbf{\omega_o}\cdot \mathbf{v} > 0$,}\\
        -(\mathbf{v} + \Delta \mathbf{n}_2)& \text{otherwise.}
    \end{cases}
\end{equation}\label{eq:normal}
To prevent the normal residual from deviating too much from the shortest axis, we add a penalty towards normal residual, making sure it is small enough.
\begin{equation}
    \mathcal{L}_{\text{reg}}=||\Delta \mathbf{n}||^2
\end{equation}
The normal residual is shown on the left of Fig.~\ref{fig:normal}.

\textbf{Normal-geometry consistency.} 
The above shortest axis direction and normal residuals are defined on each Gaussian sphere separately. However, a noticeable problem is that a normal reveals the gradient of the local geometry, which is supposed to be associated with all the Gaussian spheres in a local region. We find that simply applying a color loss to train the aforementioned normal residuals leads to inconsistency between the local geometry and the estimated normals. The main reason is that every Gaussian sphere learns its normal residuals separately without knowing the local geometry formulated by its neighbor Gaussian spheres. Thus, we have to correlate multiple Gaussian spheres in a local region with their normals to ensure normal-geometry consistency. A straightforward and naive solution is to search for its $K$ neighborhoods in space and estimate a coarse normal from all the neighboring spheres. However, such KNN searching would be extremely expensive during training because all the Gaussian spheres are dynamically moving in the optimization process. Instead, we propose a simple yet effective way to ensure normal-geometry consistency as follows.

We associate the local geometry with predicted normals by minimizing the difference between the grad normals derived from the rendered depth map and the rendered normal maps using the predicted normals
\begin{equation}
    \mathcal{L}_{\text{normal}} = ||\bar{\mathbf{n}}-\hat{\mathbf{n}}||^2,
\end{equation}
where $\bar{\mathbf{n}}$ is the rendered normal map and $\hat{\mathbf{n}}$ is computed by applying the Sobel-like operator on the rendered depth map. $\hat{\mathbf{n}}$ reveals the local geometry formulated by multiple Gaussian spheres because it is computed from the rendered depth maps. $\bar{\mathbf{n}}$ contains the information from the separately-defined normal on each Gaussian sphere. By minimizing the difference, we enforce the consistency between the local geometry and the estimated normals.


\subsection{Losses}\label{sec_loss}

In addition to $\mathcal{L}_{\text{color}}$, $\mathcal{L}_{reg}$, and $\mathcal{L}_\text{normal}$, we make use of sparse loss~\cite{lombardi2019neural, xu2022point} to encourages Gaussian spheres' opacity values $\alpha$ to approach either 0 or 1 by
\begin{equation}
    \mathcal{L}_{\text{sparse}} = \frac{1}{|\alpha|} \sum_{\alpha_i} \left[ \log(\alpha_i) + \log(1 - \alpha_i) \right].
\end{equation}
This sparsity loss helps the geometry of Gaussian spheres converge to a single thin plate and improves the rendering quality.
In summary, the total training loss $\mathcal{L}$ is 
\begin{equation}
    \mathcal{L} = 
    \mathcal{L}_{\text{color}} + 
    \lambda_\text{n}\mathcal{L}_\text{normal} + \lambda_\text{s}\mathcal{L}_\text{sparse} + 
    \lambda_\text{r}\mathcal{L}_\text{reg} 
\end{equation}
where $\lambda_\text{n}=0.01, \lambda_\text{s}=0.001, \lambda_\text{r}=0.001$.

%% file: sec/4_experiment.tex
\begin{table*}
\centering
\caption{The quantitative comparisons (PSNR / SSIM / LPIPS) on NeRF Synthetic dataset~\cite{mildenhall2021nerf}.}
\vspace{-2ex}
\scalebox{0.9}{
\begin{tabular}{lccccccccc}
\hline
\multicolumn{10}{c}{NeRF Synthetic \cite{mildenhall2021nerf}}  \\
            & \multicolumn{1}{l}{Chair} & \multicolumn{1}{l}{Drums} & \multicolumn{1}{l}{Lego} & \multicolumn{1}{l}{Mic} & \multicolumn{1}{l}{Materials} & \multicolumn{1}{l}{Ship} & \multicolumn{1}{l}{Hotdog} & \multicolumn{1}{l}{Ficus} & \multicolumn{1}{l}{Avg.}           \\ \hline
\multicolumn{10}{c}{PSNR$\uparrow$}                                                                                                                                                                                                                                               \\ \hline
NeRF~\cite{mildenhall2021nerf}        & 33.00                     & 25.01                     & 32.54                    & 32.91                   & \cellcolor[RGB]{255, 248, 174}29.62                         & 28.65                    & 36.18                      & \cellcolor[RGB]{255, 248, 174}30.13   &  31.01                          \\
VolSDF~\cite{yariv2021volume}        & 30.57                     & 20.43                     & 29.46                    & 30.53                  & 29.13                         & 25.51                    & 35.11                      & 22.91 & 27.96                             \\
Ref-NeRF~\cite{verbin2022ref}    & \cellcolor[RGB]{255, 248, 174}33.98                     & \cellcolor[RGB]{255, 248, 174}25.43                     & \cellcolor[RGB]{255, 248, 174}35.10                    & \cellcolor[RGB]{255, 248, 174}33.65                   & 27.10                         & \cellcolor[RGB]{255, 248, 174}29.24                   & \cellcolor[RGB]{255, 248, 174}37.04                      & 28.74    &  \cellcolor[RGB]{255, 248, 174}31.29                        \\
ENVIDR~\cite{liang2023envidr}    &  31.22                    & 22.99                    &   29.55                 &  32.17                  &   29.52                      &  21.57                   &    31.44                  &   26.60  &  28.13                                                                                                                                                                                                                                                                     \\ \hline
Gaussian Splatting~\cite{kerbl20233d}  &  \cellcolor[RGB]{255, 204, 153}35.82                    &  \cellcolor[RGB]{255, 204, 153}26.17                    & \cellcolor[RGB]{255, 204, 153}35.69                   &   \cellcolor[RGB]{255, 153, 153}35.34                &  \cellcolor[RGB]{255, 204, 153}30.00                        & \cellcolor[RGB]{255, 153, 153}30.87                   & \cellcolor[RGB]{255, 204, 153}37.67                      & \cellcolor[RGB]{255, 204, 153}34.83  & \cellcolor[RGB]{255, 204, 153}33.30                           \\
Ours & \cellcolor[RGB]{255, 153, 153}35.83                          & \cellcolor[RGB]{255, 153, 153}26.36                          & \cellcolor[RGB]{255, 153, 153}35.87                         & \cellcolor[RGB]{255, 204, 153}35.23                        & \cellcolor[RGB]{255, 153, 153}30.07                             & \cellcolor[RGB]{255, 204, 153}30.82                       & \cellcolor[RGB]{255, 153, 153}37.85                           & \cellcolor[RGB]{255, 153, 153}34.97  &  \cellcolor[RGB]{255, 153, 153}33.38                          \\ \hline \hline
\multicolumn{10}{c}{SSIM$\uparrow$}                                                                                                                                                                                                                                            \\ \hline
NeRF~\cite{mildenhall2021nerf}        & 0.967                     & 0.925                     & 0.961                    & 0.980                   & 0.949                         & 0.856                    & \cellcolor[RGB]{255, 248, 174}0.974                      & \cellcolor[RGB]{255, 248, 174}0.964  & 0.947                          \\
VolSDF~\cite{yariv2021volume}        & 0.949                           & 0.893                          &  0.951                        &  0.969                       &  \cellcolor[RGB]{255, 248, 174}0.954                             &   0.842                        &   0.972                         &  0.929 &   0.932                             \\
Ref-NeRF~\cite{verbin2022ref}    & \cellcolor[RGB]{255, 248, 174}0.974                     & 0.929                     & \cellcolor[RGB]{255, 204, 153}0.975                    & \cellcolor[RGB]{255, 248, 174}0.983                   & 0.921                         & \cellcolor[RGB]{255, 248, 174}0.864                    & \cellcolor[RGB]{255, 204, 153}0.979                      & 0.954 & 0.947                             \\
ENVIDR~\cite{liang2023envidr}    &  \cellcolor[RGB]{255, 204, 153}0.976                    & \cellcolor[RGB]{255, 248, 174}0.930                    &   \cellcolor[RGB]{255, 248, 174}0.961                 &  \cellcolor[RGB]{255, 204, 153}0.984                  & \cellcolor[RGB]{255, 153, 153}0.968                        &   0.855                  &  0.963                    & \cellcolor[RGB]{255, 153, 153}0.987  &  \cellcolor[RGB]{255, 248, 174}0.956                                                                                                                                                                                                                                                                     \\ \hline
Gaussian Splatting~\cite{kerbl20233d}  &     \cellcolor[RGB]{255, 153, 153}0.987                 &   \cellcolor[RGB]{255, 153, 153}0.954                   &  \cellcolor[RGB]{255, 153, 153}0.983                   & \cellcolor[RGB]{255, 153, 153}0.991                   &  \cellcolor[RGB]{255, 204, 153}0.960                        &  \cellcolor[RGB]{255, 153, 153}0.907                   &   \cellcolor[RGB]{255, 153, 153}0.985                    &  \cellcolor[RGB]{255, 153, 153}0.987  &  \cellcolor[RGB]{255, 153, 153}0.969                        \\
Ours &           \cellcolor[RGB]{255, 153, 153}0.987                &     \cellcolor[RGB]{255, 204, 153}0.949                      &           \cellcolor[RGB]{255, 153, 153}0.983               &       \cellcolor[RGB]{255, 153, 153}0.991                  &             \cellcolor[RGB]{255, 204, 153}0.960                  &        \cellcolor[RGB]{255, 204, 153}0.905                  &          \cellcolor[RGB]{255, 153, 153}0.985                  &            \cellcolor[RGB]{255, 204, 153}0.985      &  \cellcolor[RGB]{255, 204, 153}0.968                 \\ \hline \hline
\multicolumn{10}{c}{LPIPS$\downarrow$}                                                                                                                                                                                                                                              \\ \hline
NeRF~\cite{mildenhall2021nerf}        & 0.046                     & 0.091                     & 0.050                    & 0.028                   & 0.063                         & 0.206                    & 0.121                      & 0.044  & 0.081                        \\
VolSDF~\cite{yariv2021volume}        &  0.056                          & 0.119                          &   0.054                        &   0.191                       &  0.048                             &   0.191                        &   0.043                          & 0.068    &   0.096                             \\
Ref-NeRF~\cite{verbin2022ref}    & \cellcolor[RGB]{255, 204, 153}0.029                     & \cellcolor[RGB]{255, 248, 174}0.073                     & \cellcolor[RGB]{255, 248, 174}0.025                    & \cellcolor[RGB]{255, 204, 153}0.018                   & 0.078                         & \cellcolor[RGB]{255, 248, 174}0.158                    & \cellcolor[RGB]{255, 248, 174}0.028                      & 0.056   & \cellcolor[RGB]{255, 248, 174}0.058                        \\
ENVIDR~\cite{liang2023envidr}    &  \cellcolor[RGB]{255, 248, 174}0.031                    &  0.080                   &  0.054                  &       \cellcolor[RGB]{255, 248, 174}0.021             & \cellcolor[RGB]{255, 248, 174}0.045                        &   0.228                  &  0.072                    &  \cellcolor[RGB]{255, 153, 153}0.010    &  0.067                                                                                                                                                                                                                                                                     \\ \hline
Gaussian Splatting~\cite{kerbl20233d}  &  \cellcolor[RGB]{255, 153, 153}0.012                    &  \cellcolor[RGB]{255, 153, 153}0.037                    &  \cellcolor[RGB]{255, 204, 153}0.016                   & \cellcolor[RGB]{255, 153, 153}0.006                   &  \cellcolor[RGB]{255, 204, 153}0.034                        &  \cellcolor[RGB]{255, 204, 153}0.106                   &  \cellcolor[RGB]{255, 204, 153}0.020                     &  \cellcolor[RGB]{255, 204, 153}0.012  & \cellcolor[RGB]{255, 204, 153}0.030                          \\
Ours &        \cellcolor[RGB]{255, 153, 153}0.012                   &          \cellcolor[RGB]{255, 204, 153}0.040                 &             \cellcolor[RGB]{255, 153, 153}0.014             &         \cellcolor[RGB]{255, 153, 153}0.006                &              \cellcolor[RGB]{255, 153, 153}0.033                 &          \cellcolor[RGB]{255, 153, 153}0.098                &             \cellcolor[RGB]{255, 153, 153}0.019                &        \cellcolor[RGB]{255, 248, 174}0.013        &  \cellcolor[RGB]{255, 153, 153}0.029            \\ \hline \hline
\end{tabular}}
\label{tab:tab_nerfsynthetic}
\end{table*}

\begin{figure*}[t]
    \centering
    \includegraphics[width=\linewidth]{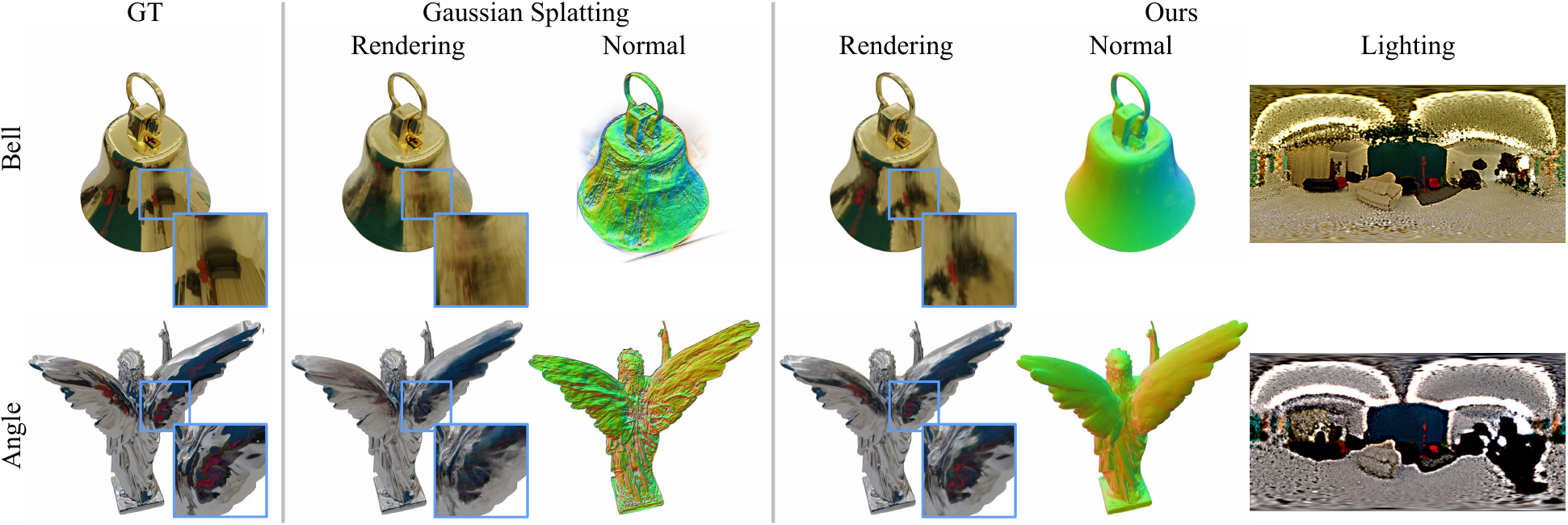}
    \caption{The qualitative comparisons with 3D Gaussian Splatting~\cite{kerbl20233d} on Glossy dataset~\cite{liu2023nero}. Our method not only renders the objects with high fidelity but also provides detailed normal and lighting maps. Some areas are zoomed in for better visualization.
    }
    \label{fig:glossy_synthetic}
      \vspace{-3ex}
\end{figure*}

\section{Experiments}
\subsection{Datasets}
To comprehensively validate the effectiveness of GaussianShader, we conduct evaluation on various datasets:  \textbf{a)} widely-used NVS dataset: NeRF Synthetic~\cite{mildenhall2021nerf}. \textbf{b)} reflective objects datasets: Shiny Blender~\cite{verbin2022ref} and Glossy Synthetic~\cite{liu2023nero}. \textbf{c)} real-world large-scale scenes: Tanks and Temples~\cite{knapitsch2017tanks}. 

\subsection{Baselines and Metrics}

We compare our method against the following baselines: 
\textbf{a)} 3D Gaussian Splatting~\cite{kerbl20233d}: a real-time radiance field rendering method based on efficient 3D Gaussian representation;
\textbf{b)} VolSDF~\cite{yariv2021volume}: a classical neural implicit surface reconstruction method based on SDF;  \textbf{c)} Ref-NeRF~\cite{verbin2022ref}: state-of-the-art method in novel view synthesis of reflective objects; \textbf{d)} NVDiffRec~\cite{munkberg2022extracting}, NVDiffRecMC~\cite{hasselgren2022nvdiffrecmc}: top-performing neural inverse rendering methods; \textbf{e)} ENVIDR~\cite{liang2023envidr}, NeRO~\cite{liu2023nero}: top-performing SDF-based neural implicit methods for reconstructing reflective objects.
The evaluation metrics to measure rendering quality are reported in PSNR, SSIM~\cite{wang2004image}, and LPIPS~\cite{zhang2018unreasonable}.

\subsection{Implementation Details}
All experiments are conducted on a Nvidia RTX 3090 graphics card. We optimize our models using Adam~\cite{kingma2014adam} optimizer for 30,000 iterations. 
For Ref-NeRF~\cite{verbin2022ref}, ENVIDR~\cite{liang2023envidr}, and 3D Gaussian Splatting~\cite{kerbl20233d}, we retrain them using their official codes and obtain all their results from our retrained models. Other works' results are imported from their original papers. 
\begin{figure}[t]
    \centering
    \includegraphics[width=1\linewidth]{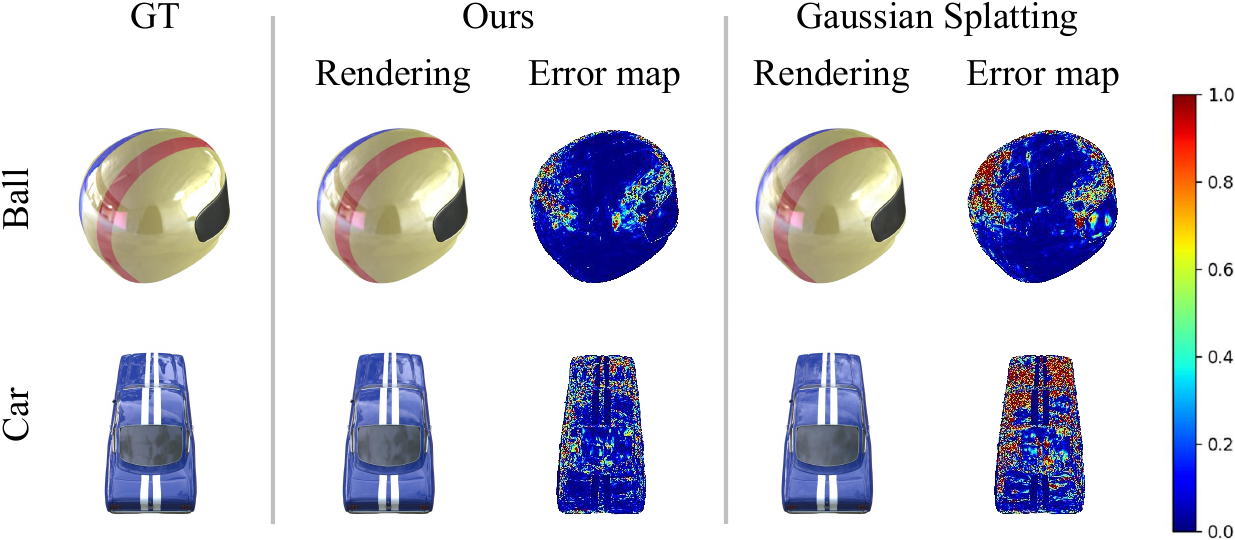}
    \caption{Qualitative comparisons on Shiny Blender dataset~\cite{verbin2022ref}. Error maps corresponding to each rendering result demonstrate the enhanced fidelity of our approach in highly specular areas.}
    \label{fig:fig_shinyblender_rgb}
\end{figure}
\subsection{Comparisons}

\begin{figure*}
    \centering
    \includegraphics[width=\linewidth]{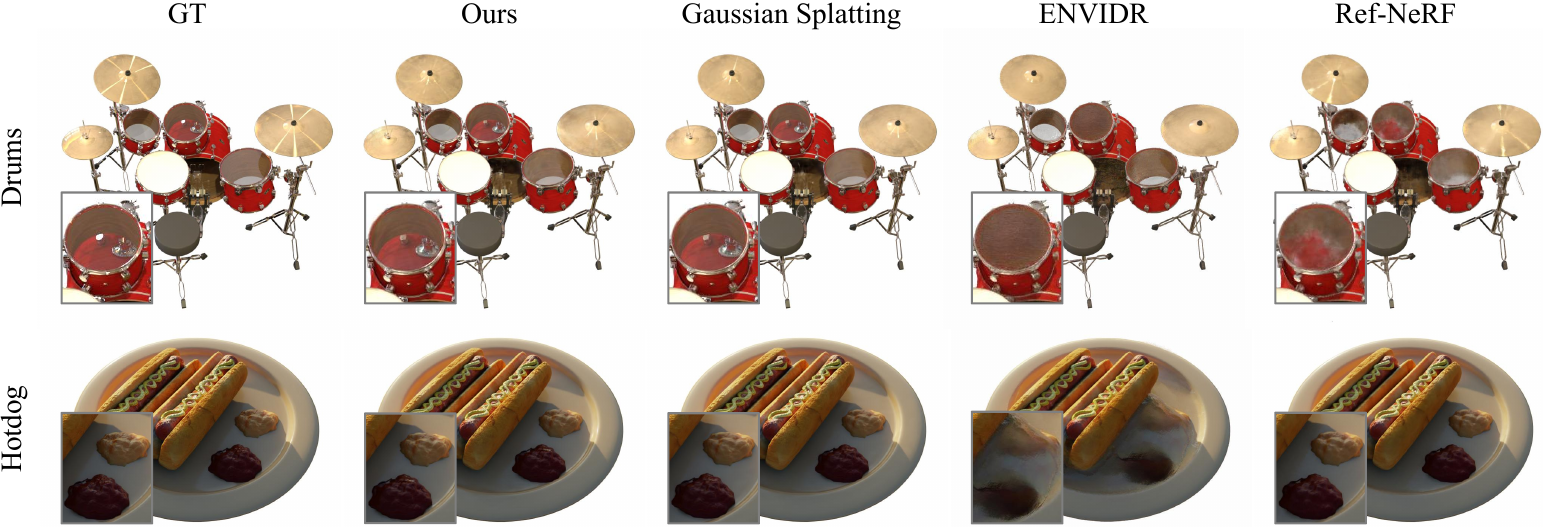}
    \caption{
   We present a qualitative comparisons of our method on NeRF Synthetic dataset~\cite{mildenhall2021nerf} against previous techniques and corresponding ground truth images from test views. For clarity, we zoom in the images where differences in quality are especially notable.
    }
    \label{fig:fig_nerfsynthetic_rgb}
\end{figure*}

\textbf{NeRF Synthetic dataset~\cite{mildenhall2021nerf}.} We first evaluate our model on the NeRF Synthetic dataset, which contains objects with complex geometry and realistic non-Lambertian materials. We show the quantitative results in Tab.~\ref{tab:tab_nerfsynthetic} and visual comparisons in Fig.~\ref{fig:fig_nerfsynthetic_rgb}. Our approach achieves numerically and visually comparable results with both Gaussian Splatting and neural rendering methods, demonstrating the effectiveness of our method in rendering general objects. Note that SDF-based ENVIDR~\cite{liang2023envidr} cannot perform well on shadowed regions of complex objects, as shown in Fig.~\ref{fig:fig_nerfsynthetic_rgb}.

\begin{figure}
    \centering
    \includegraphics[width=\linewidth]{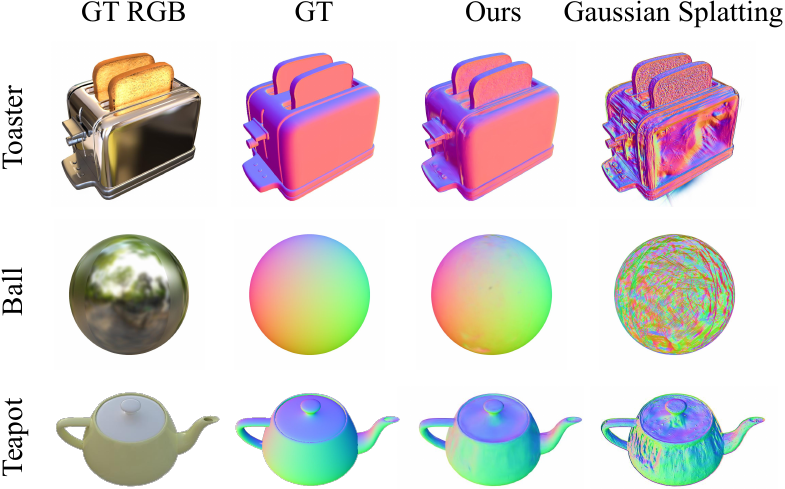}
    \caption{Visualized normal results on Shiny Blender dataset~\cite{verbin2022ref}, compared with Gaussian Splatting~\cite{kerbl20233d}. It illustrates the superior normal estimation achieved by our method. 
    }
    \label{fig:fig_shinyblender_normal}
          \vspace{-3ex}
\end{figure}

\textbf{Shiny Blender dataset~\cite{verbin2022ref}.} 
Quantitative results on the Shiny Blender dataset are reported in Tab.~\ref{tab:tab_shinyblender}. With the ability to model light-surface interactions, our method outperforms original Gaussian Splatting in all scenes while slightly underperforms ENVIDR and Ref-NeRF. ENVIDR capitalizes on the continuous properties of Signed Distance Fields to naturally create smoother surfaces and normals, resulting in high rendering quality, particularly for reflective objects. Fig.~\ref{fig:fig_shinyblender_rgb} shows our comparisons with Gaussian Splatting. We can see that our method correctly renders surfaces with strong specular appearances. High-quality modeling of reflections is contingent upon the accurate estimation of normals as shown in Fig.~\ref{fig:fig_shinyblender_normal}.

Additionally, we report the average PSNR, training time, and rendering FPS in Tab.~\ref{tab:tab_speed}. The reported results are averaged among all objects of the Shiny Blender and NeRF synthetic datasets. On the same hardware, our method only takes about 0.5 hour for training while MLP-based methods, like ENVIDR and Ref-NeRF, require 6 hours and 23 hours for optimizing. 
Due to the introduce of extra lighting and shading attributes, our method is a bit slower than 3D Gaussian Splatting~\cite{kerbl20233d} but still keeps reasonable efficiency and real-time rendering speed.


\begin{table}
\centering
\caption{Quantative comparisons on Shiny Blender dataset~\cite{verbin2022ref}. Our method is comparable with both Gaussian Splatting~\cite{kerbl20233d} and prior reflective object reconstruction methods.}
\vspace{-1ex}
\scalebox{0.63}{
\begin{tabular}{lccccccc}
\hline
\multicolumn{8}{c}{Shiny Blender~\cite{verbin2022ref}}                                                                                                                                                   \\
         & \multicolumn{1}{l}{Car} & \multicolumn{1}{l}{Ball} & \multicolumn{1}{l}{Helmet} & \multicolumn{1}{l}{Teapot} & \multicolumn{1}{l}{Toaster} & \multicolumn{1}{l}{Coffee} & \multicolumn{1}{l}{Avg.}\\ \hline
\multicolumn{8}{c}{PSNR$\uparrow$}                                                                                                                                                           \\ \hline
NVDiffRec~\cite{munkberg2022extracting} & \cellcolor[RGB]{255, 248, 174}27.98  & 21.77 &  26.97  & 40.44 &  24.31 & 30.74 & 28.70 \\
NVDiffMC~\cite{hasselgren2022nvdiffrecmc} &25.93  &  \cellcolor[RGB]{255, 248, 174}30.85  & 26.27  & 38.44 &  22.18 &  29.60 & 28.88\\
Ref-NeRF~\cite{verbin2022ref} & \cellcolor[RGB]{255, 153, 153}30.41                   & 29.14                    & \cellcolor[RGB]{255, 204, 153}29.92                      & \cellcolor[RGB]{255, 248, 174}45.19                      & 25.29                       & \cellcolor[RGB]{255, 153, 153}33.99   & \cellcolor[RGB]{255, 204, 153}32.32                  \\
NeRO~\cite{liu2023nero}     & 25.53                   & 30.26                        & \cellcolor[RGB]{255, 248, 174}29.20                          & 38.70                           &   \cellcolor[RGB]{255, 153, 153}26.46                          & 28.89       &  29.84                  \\
ENVIDR~\cite{liang2023envidr}   & \cellcolor[RGB]{255, 204, 153}28.46                   & \cellcolor[RGB]{255, 153, 153}38.89                    & \cellcolor[RGB]{255, 153, 153}32.73                      & 41.59                      & \cellcolor[RGB]{255, 248, 174}26.11                       & 29.48    & \cellcolor[RGB]{255, 153, 153}32.88                 \\ \hline
Guassian Splatting~\cite{kerbl20233d} & 27.24 & 27.69 & 28.32 & \cellcolor[RGB]{255, 204, 153}45.68 & 20.99 & \cellcolor[RGB]{255, 248, 174}32.32 & 30.37 \\
Ours     &      27.90                   &       \cellcolor[RGB]{255, 204, 153}30.98                   &         28.32                   &         \cellcolor[RGB]{255, 153, 153}45.86                   &                \cellcolor[RGB]{255, 204, 153}26.21             &            \cellcolor[RGB]{255, 204, 153}32.39   &  \cellcolor[RGB]{255, 248, 174}31.94           \\ \hline \hline
\multicolumn{8}{c}{SSIM$\uparrow$}                                                                                                                                                           \\ \hline
NVDiffRec~\cite{munkberg2022extracting} & \cellcolor[RGB]{255, 153, 153}0.963 &  0.858 &  0.951 &  \cellcolor[RGB]{255, 153, 153}0.996 &  \cellcolor[RGB]{255, 248, 174}0.928 &  \cellcolor[RGB]{255, 153, 153}0.973 & 0.945 \\
NVDiffMC~\cite{hasselgren2022nvdiffrecmc} & 0.940 &  0.940 &  0.940 &  \cellcolor[RGB]{255, 204, 153}0.995 & 0.886  &  0.965 & 0.944 \\
Ref-NeRF~\cite{verbin2022ref} & \cellcolor[RGB]{255, 248, 174}0.949                   & 0.956                    & \cellcolor[RGB]{255, 248, 174}0.955                      & \cellcolor[RGB]{255, 204, 153}0.995                      & 0.910                       & \cellcolor[RGB]{255, 204, 153}0.972   & 0.956                  \\
NeRO~\cite{liu2023nero}     & \cellcolor[RGB]{255, 248, 174}0.949                    & \cellcolor[RGB]{255, 204, 153}0.974                         & \cellcolor[RGB]{255, 204, 153}0.971                           & \cellcolor[RGB]{255, 204, 153}0.995                           & \cellcolor[RGB]{255, 204, 153}0.929                            & 0.956   & \cellcolor[RGB]{255, 204, 153}0.962                       \\
ENVIDR~\cite{liang2023envidr}   & \cellcolor[RGB]{255, 204, 153}0.961                   & \cellcolor[RGB]{255, 153, 153}0.991                    & \cellcolor[RGB]{255, 153, 153}0.980                      & \cellcolor[RGB]{255, 153, 153}0.996                      & \cellcolor[RGB]{255, 153, 153}0.939                       & 0.949    &   \cellcolor[RGB]{255, 153, 153}0.969               \\ \hline
Guassian Splatting~\cite{kerbl20233d} & 0.930 & 0.937 & 0.951 & \cellcolor[RGB]{255, 153, 153}0.996 & 0.895 & \cellcolor[RGB]{255, 248, 174}0.971 & 0.947 \\
Ours     &    0.931                     &           \cellcolor[RGB]{255, 248, 174}0.965               &       0.950                     &        \cellcolor[RGB]{255, 153, 153}0.996                    &          \cellcolor[RGB]{255, 204, 153}0.929                   &       \cellcolor[RGB]{255, 248, 174}0.971   &  \cellcolor[RGB]{255, 248, 174}0.957                \\ \hline \hline
\multicolumn{8}{c}{LPIPS$\downarrow$}                                                                                                                                                          \\ \hline
NVDiffRec~\cite{munkberg2022extracting} & \cellcolor[RGB]{255, 153, 153}0.045 &  0.297 &  0.118 &  \cellcolor[RGB]{255, 204, 153}0.011 &  0.169 &  \cellcolor[RGB]{255, 153, 153}0.076 & 0.119 \\
NVDiffMC~\cite{hasselgren2022nvdiffrecmc} & 0.077 &  0.312 &  0.157 &  0.014 &  0.225 &  0.097 & 0.147 \\
Ref-NeRF~\cite{verbin2022ref} & 0.051                   & 0.307                    & 0.087                      & 0.013                      & 0.118                       & \cellcolor[RGB]{255, 248, 174}0.082  & 0.109                   \\
NeRO~\cite{liu2023nero}     & 0.074 & \cellcolor[RGB]{255, 204, 153}0.094 & \cellcolor[RGB]{255, 153, 153}0.050   & \cellcolor[RGB]{255, 248, 174}0.012  & \cellcolor[RGB]{255, 204, 153}0.089 & 0.110      & \cellcolor[RGB]{255, 204, 153}0.072                     \\
ENVIDR~\cite{liang2023envidr}   & \cellcolor[RGB]{255, 248, 174}0.049                   & \cellcolor[RGB]{255, 153, 153}0.067                    & \cellcolor[RGB]{255, 204, 153}0.051                      & \cellcolor[RGB]{255, 204, 153}0.011                      & \cellcolor[RGB]{255, 248, 174}0.116                       & 0.139   &  \cellcolor[RGB]{255, 204, 153}0.072                 \\ \hline
Guassian Splatting~\cite{kerbl20233d} & \cellcolor[RGB]{255, 204, 153}0.047 & 0.161 & 0.079 & \cellcolor[RGB]{255, 153, 153}0.007 & 0.126 & \cellcolor[RGB]{255, 204, 153}0.078 & \cellcolor[RGB]{255, 248, 174}0.083\\
Ours     &         \cellcolor[RGB]{255, 153, 153}0.045                &            \cellcolor[RGB]{255, 248, 174}0.121              &          \cellcolor[RGB]{255, 248, 174}0.076                  &        \cellcolor[RGB]{255, 153, 153}0.007                    &              \cellcolor[RGB]{255, 153, 153}0.079               &            \cellcolor[RGB]{255, 204, 153}0.078  &  \cellcolor[RGB]{255, 153, 153}0.068            \\ \hline \hline
\end{tabular}
}
\label{tab:tab_shinyblender}
      \vspace{-2ex}
\end{table}

\begin{figure}[h]
    \centering
    \includegraphics[width=\linewidth]{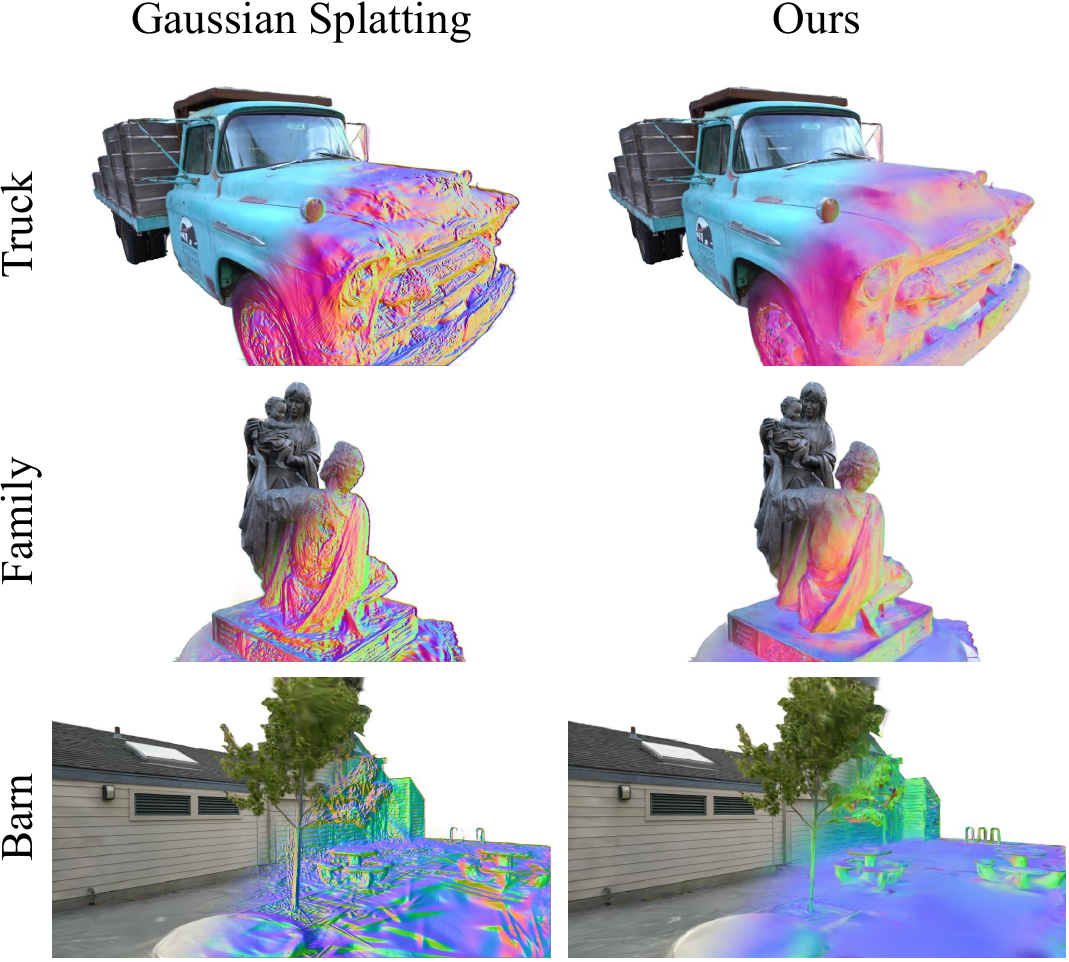}
    \caption{Qualitative comparisons with Gaussian Splatting~\cite{kerbl20233d} on the Tanks and Temples dataset~\cite{knapitsch2017tanks}. Our approach yields a smoother and more reasonable normal.}
    \label{fig:tanks_normal}
      \vspace{-3ex}
\end{figure}

\begin{table}[]
\centering
\caption{Comparisons over average PSNR, training time and FPS. Our method achieves the best rendering quality and competitive training/rendering speed.}
\vspace{-1ex}
\scalebox{0.8}{
\begin{tabular}{lccc}
\hline
                   & PSNR  & Training Time             & FPS            \\ \hline
Ref-NeRF~\cite{verbin2022ref}            & 31.73 & 23h & 0.03 \\
ENVIDR~\cite{liang2023envidr}              & 30.04 & 6h                & 1.33           \\
Gaussian Splatting~\cite{kerbl20233d} & 32.05 & \textbf{0.25h}            & \textbf{274}            \\
Ours               & \textbf{32.76} & 0.58h            & 97             \\ \hline
\end{tabular}}
\label{tab:tab_speed}
\end{table}

\textbf{Tanks and Temples dataset~\cite{knapitsch2017tanks}.} To explore our method's scalability in larger-scale environments instead of only small objects, we conducted experiments using Tanks and Temples \cite{knapitsch2017tanks}. The qualitative results in Fig.~\ref{fig:tanks_normal} show that our model outperforms Gaussian Splatting~\cite{kerbl20233d}. 

\textbf{Glossy Synthetic dataset~\cite{liu2023nero}.} We present a comparative visualization between Gaussian Splatting and our method on the Glossy dataset shown in Fig.~\ref{fig:glossy_synthetic}. Our method not only renders objects with exceptional fidelity but also provides more accurate normal and lightings, capturing the nuances of reflective surfaces with greater precision. These maps reveal our method's ability to approximate the rendering function, resulting in a more lifelike portrayal of specular lighting. Please refer to the supplementary materials provided for an extensive showcase of our results.
\subsection{Ablation Study}
In this section, we conduct ablation studies of our model on Shiny Blender dataset. The numerical results are reported in Tab.~\ref{tab:tab_ablation}. All ablation experiments are done on Shiny Blender dataset with $1/2$ image resolution.

\textbf{Loss functions.} We trained our model serperately under the setting ``without sparse loss $\mathcal{L}_\text{sparse}$" and ``without normal loss $\mathcal{L}_\text{normal}$". 
The incorporation of these losses, especially the direct regularization on normals, proved critical, yielding a marked improvement in rendering quality by effectively guiding the normal optimization process as shown in Tab.~\ref{tab:tab_ablation}.

\textbf{Residual color.} 
Residual color, comprising 3rd order SH coefficients, is used to compensate for indirect light. As shown in Table~\ref{tab:tab_ablation},
the introduce of residual color enhances our model, allowing it to simulate a wide array of real-world intricate lighting situations and significantly improve rendering quality.

\begin{table}[]
\centering
\caption{Ablation studies on loss function, lighting representation, normal axis and residual color using Shiny Blender dataset. }
\scalebox{0.9}{
\begin{tabular}{lccc}
\hline
                                           & PSNR$\uparrow$           & SSIM$\uparrow$           & LPIPS$\downarrow$  
                                            \\ \hline
w/o $\mathcal{L}_\text{sparse}$                            & 31.79          & 0.952          & 0.056          \\
w/o $\mathcal{L}_\text{normal}$                            & 30.93          & 0.941          & 0.060           \\
w/o $\mathbf{c}_r$                & 31.49          & 0.948          & 0.060           \\
w/o $\mathbf{v}$                  & 31.47         & 0.951          & 0.058        \\
MLP Lighting                   & 29.73          & 0.936         & 0.075    \\
Full Model & \textbf{32.09} & \textbf{0.953} & \textbf{0.054} \\ \hline
\end{tabular}
}
\label{tab:tab_ablation}
      \vspace{-3ex}
\end{table}

\textbf{Normal formulation.} 
In our full model, we use the shortest axis direction $\mathbf{v}$ with a learnable residual $\Delta \mathbf{n}$ to formulate normal, as described by Eqn.~\ref{eq:normal}. In Tab.~\ref{tab:tab_ablation}, we compared this novel normal formation with a naive normal prediction manner (indicated by ``w/o axis $\mathbf{v}$"). Instead of correlating with the geometry of 3D gaussians, we directly optimize a single normal attribute.
The result shows our full model can achieve higher rendering quality, since our predicted normals are correlated with the true geometry of Gaussian spheres. 

\textbf{Lighting representation.} A crucial component of our approach is the modeling of lighting. ``MLP Lighting" means we use a directional MLP similar to Ref-NeRF to implicitly optimize lighting. In our full model, we use ``Envmap Lighting" representation as illustrated in Sec.~\ref{sec_light}. The result proves that our explicit lighting representation is better suited for 3D Gaussian rendering framework to correctly render reflective objects. 

%% file: sec/5_conclusion.tex
\section{Conclusions}
In summary, we propose GaussianShader, advancing the rendering of both reflective and general objects through an extended 3D Gaussian model. 
Specifically, our method integrates shading functions with 3D Gaussian representation for handling view-dependent appearances, successfully handling reflective surfaces.
Moreover, a novel normal prediction method is proposed to achieve high-quality normals, which correlates the shading parameters and true geometry of 3D Gaussians.
The results show our approach achieves high-quality renderings, marking a substantial improvement in efficient, realistic 3D object rendering.


%% file: sec/X_suppl.tex
\clearpage
\setcounter{page}{1}
\maketitlesupplementary

\section{Residual Color Training Details}
Residual color, comprising 3rd order SH coefficients, 
is initially excluded to ensure color learning is attributed to materials and lighting for accurate material separation, and introduced in the later training stages for refinement.

\section{Additional Results}
\begin{figure*}[h]
    \centering
    \includegraphics[width=0.9\linewidth]{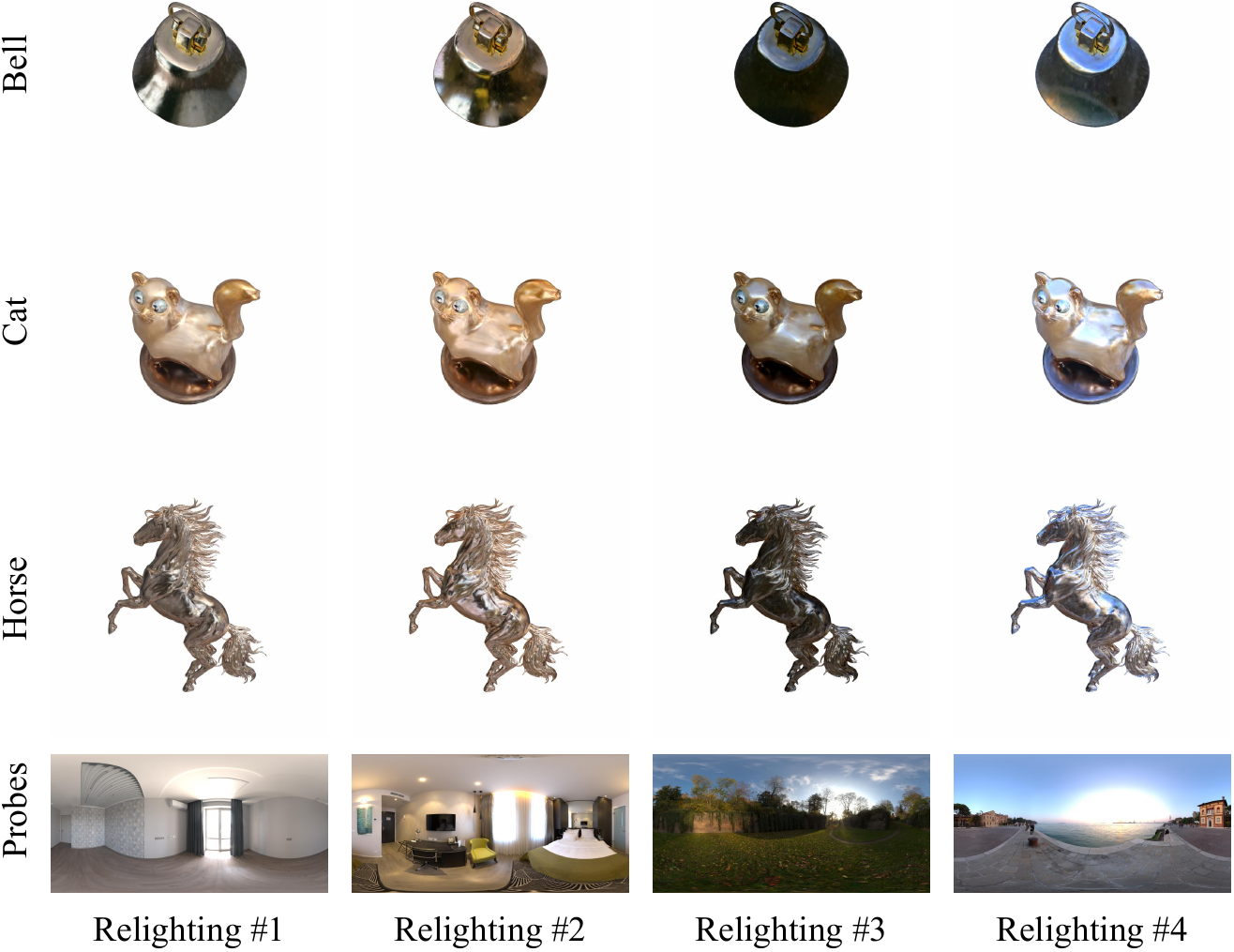}
    \caption{The image showcases a series of relighting results, where objects are from Glossy Synthetic~\cite{liu2023nero} and rendered under four distinct lighting environments. Each column represents a different lighting condition, highlighting our method's ability to adaptively recast objects in diverse and dynamic light settings, from indoor ambient to outdoor natural light, thus validating the robustness and flexibility of our relighting technique.}
    \label{fig:relit}
    \vspace{-3ex}
\end{figure*}

\textbf{Relighting.} 
In Fig.~\ref{fig:relit}, we display objects reconstructed using our method, including relighting scenarios with different lighting conditions, featuring both warm and cool tones, and indoor and outdoor environments. Our renderings, spanning four diverse sets, convincingly show that the relit scenes maintain realism, with object highlights effectively mirroring the surrounding light sources, exemplifying our method's proficiency in relighting applications.

\textbf{Quantiative results on Glossy Synthetic~\cite{liu2023nero}.}
Compared to Gaussian Splatting~\cite{kerbl20233d}, our method registers a significant improvement on the Glossy Synthetic dataset~\cite{liu2023nero}, boasting an average PSNR increase of 1.1 as shown in Tab.~\ref{tab:tab_glossy}. This superior performance stems from our detailed modeling of lighting and shading attributes, which facilitates a more accurate depiction of how light interacts with various surfaces. Precise normal orientation further refines this interaction, leading to renderings that faithfully reproduce the subtle reflections and refractions typical of glossy materials, thereby enhancing overall scene fidelity.

\textbf{Quantiative results on Tanks and Temples~\cite{knapitsch2017tanks}.}
In Tab.~\ref{tab:tab_tanks}, our method shows a modest improvement on the Tanks and Temples dataset~\cite{knapitsch2017tanks} comparing with Gaussian Splatting\cite{kerbl20233d}. This is primarily because the objects within this dataset are predominantly diffuse, which does not fully leverage the strengths of our approach that are more pronounced in handling complex lighting scenarios. 

\begin{table}[h]
\centering
\caption{Quantitative comparisons with Guassian Splatting~\cite{kerbl20233d} on Glossy Synthetic~\cite{liu2023nero}}
\vspace{-1ex}
\scalebox{0.67}{
\begin{tabular}{lccccccccc}
\hline
\multicolumn{10}{c}{Glossy Synthetic~\cite{liu2023nero}}                                                                                                                                                   \\
         & \multicolumn{1}{l}{Angel} & \multicolumn{1}{l}{Bell} & \multicolumn{1}{l}{Cat} & \multicolumn{1}{l}{Horse} & \multicolumn{1}{l}{Luyu} & \multicolumn{1}{l}{Potion}  & \multicolumn{1}{l}{Tbell}  & \multicolumn{1}{l}{Teapot} & \multicolumn{1}{l}{Avg.}\\ \hline
\multicolumn{10}{c}{PSNR$\uparrow$}                                                                                                                                                           \\ \hline
\cite{kerbl20233d} &26.98	&25.03	&31.15	&25.18	&26.89	&29.79	&23.92	&21.18	&26.26
 \\
Ours      &28.07	 &28.08	 &31.81	 &25.53	 &27.25	 &30.08	 &24.48	 &23.57	 &27.36
           \\ \hline 
\multicolumn{10}{c}{SSIM$\uparrow$}                                                                                                                                                           \\ \hline
\cite{kerbl20233d} &  0.915	&  0.901	&  0.959	&  0.910	&  0.916	&  0.934	&  0.901	&  0.875	&  0.914
 \\
Ours    & 0.923	& 0.920	& 0.961	&0.918	& 0.915	& 0.936	& 0.897	& 0.899	& 0.921
               \\ \hline
\multicolumn{10}{c}{LPIPS$\downarrow$}                                                                                                                                                          \\ \hline
\cite{kerbl20233d} & 0.070	& 0.107	& 0.060	& 0.067	& 0.064	& 0.090	& 0.119	& 0.102	& 0.085
\\
Ours     & 0.065	& 0.097	& 0.056	& 0.062	& 0.064	& 0.087	& 0.121	& 0.090	& 0.080
          \\ \hline
\end{tabular}
}
\label{tab:tab_glossy}
\end{table}

\begin{table}
\centering
\caption{Quantitative comparisons with Guassian Splatting~\cite{kerbl20233d} on Tanks and Temples~\cite{knapitsch2017tanks}.}
\scalebox{0.67}{
\begin{tabular}{lcccccc}
\hline
\multicolumn{7}{c}{Tanks and Temples~\cite{knapitsch2017tanks}}                                                                                                                                                   \\
         & \multicolumn{1}{l}{Barn} & \multicolumn{1}{l}{Caterpillar} & \multicolumn{1}{l}{Family} & \multicolumn{1}{l}{Ignatius} & \multicolumn{1}{l}{Truck} & \multicolumn{1}{l}{Avg.}\\ \hline
\multicolumn{7}{c}{PSNR$\uparrow$}                                                                                                                                                           \\ \hline
\cite{kerbl20233d} &28.98	&26.09	&34.70	&29.52	&28.41	&29.54

 \\
Ours      &29.16	&26.19	&35.06	&29.79	&28.45	&29.73
           \\ \hline 
\multicolumn{7}{c}{SSIM$\uparrow$}                                                                                                                                                           \\ \hline
\cite{kerbl20233d} &0.921	&0.932	&0.981	&0.973	&0.945	&0.951
 \\
Ours    &0.923	&0.931	&0.982	&0.973	&0.944	&0.951
               \\ \hline
\multicolumn{7}{c}{LPIPS$\downarrow$}                                                                                                                                                          \\ \hline
\cite{kerbl20233d} &0.110	&0.075	&0.024	&0.032	&0.059	&0.060
\\
Ours     &0.104	&0.074	&0.023	&0.032	&0.056	&0.058
          \\ \hline
\end{tabular}
}
\label{tab:tab_tanks}
\end{table}